\documentclass{article} 
\usepackage{ref,times}

\usepackage{amsmath,amsfonts,bm}

\def\eqref#1{equation~\ref{#1}}

\def\1{\bm{1}}

\DeclareMathAlphabet{\mathsfit}{\encodingdefault}{\sfdefault}{m}{sl}
\SetMathAlphabet{\mathsfit}{bold}{\encodingdefault}{\sfdefault}{bx}{n}

\usepackage{hyperref}
\usepackage{url}

\usepackage{booktabs} 
\usepackage{bbm}
\usepackage{enumitem}
\usepackage{tikz}
\usepackage{xcolor}
\usepackage{wrapfig}

\usepackage[ruled]{algorithm2e} 

\SetAlFnt{\small}
\SetAlCapFnt{\small}
\SetAlCapNameFnt{\small}
\SetAlCapHSkip{0pt}

\hypersetup{
    colorlinks=true,
    linkcolor=blue,
    citecolor=blue,
    urlcolor=blue,
    filecolor=blue,
}

\def\sysname{Mono4DEditor}
\title{\sysname{}: Text-Driven 4D Scene Editing from Monocular Video via Point-Level Localization of Language-Embedded Gaussians}

\author{
Jin-Chuan Shi$^{1,2}$\thanks{Equal Contribution. Work done during M.S. study at Beihang University.}
~~
Chenye Su$^{2*}$
~~
Jiajun Wang$^{2}$
~~
Ariel Shamir$^{3}$
~~
\textbf{Miao Wang}$^{2}$\thanks{Corresponding author.}
\\[0.253cm]
$^1$ Zhejiang University, China
\\
$^2$ State Key Laboratory of Virtual Reality Technology and Systems, Beihang University, China
\\
$^3$ Reichman University, Israel
}

\iclrfinalcopy 

\begin{document}

\maketitle

\vspace{-8mm}
\begin{figure*}[h]
  \centering
  \includegraphics[width=\textwidth]{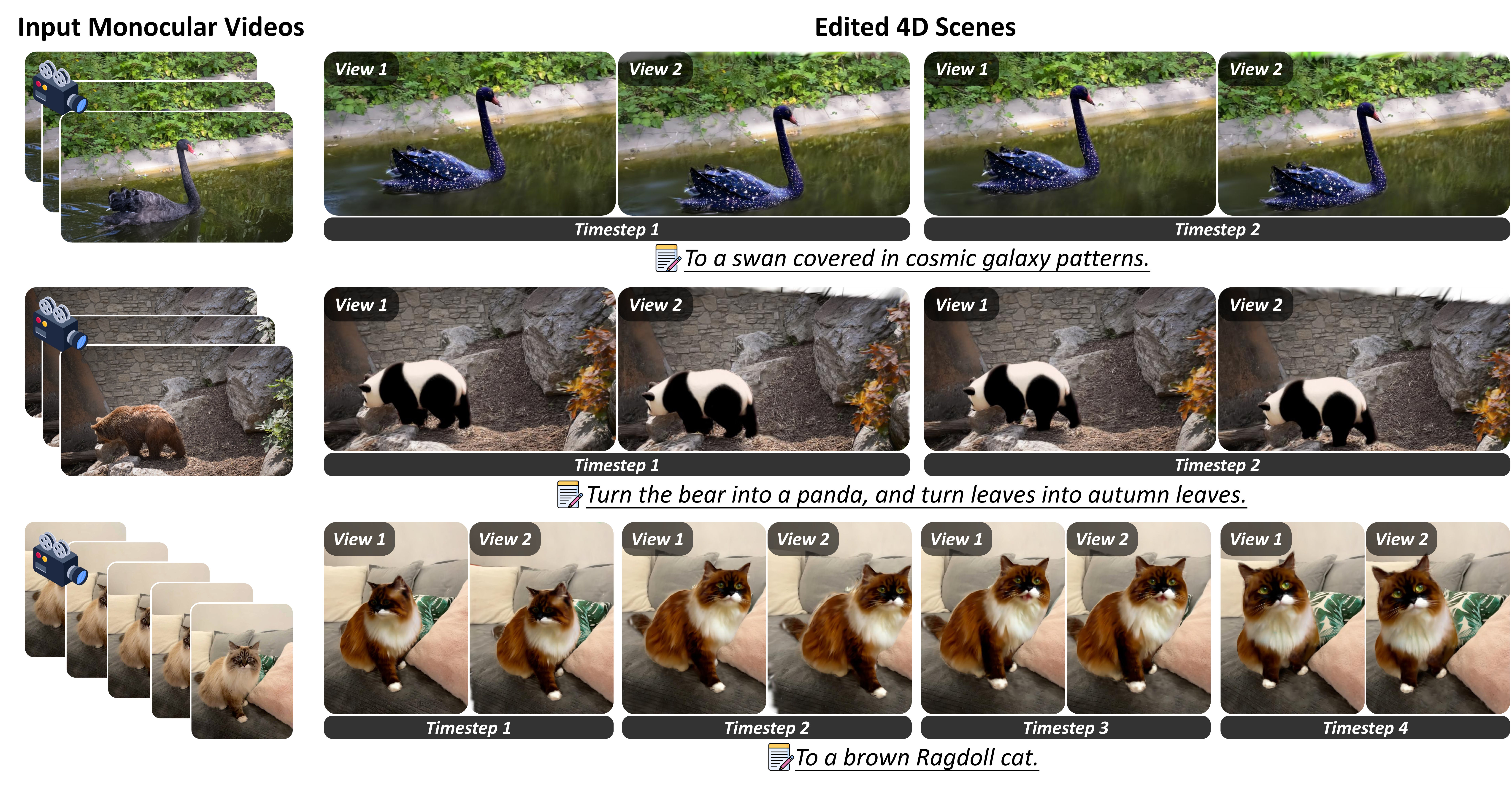}
  \caption{Our approach \sysname{} allows users to edit 4D scenes from casual monocular video with text instruction. \sysname{} achieves precise, high-quality editing of the instructed content while maintaining irrelevant regions unchanged.}
  \label{fig:teaser}
\end{figure*}

\begin{abstract}
Editing 4D scenes reconstructed from monocular videos based on text prompts is a valuable yet challenging task with broad applications in content creation and virtual environments. The key difficulty lies in achieving semantically precise edits in localized regions of complex, dynamic scenes, while preserving the integrity of unedited content. 
To address this, we introduce \sysname{}, a novel framework for flexible and accurate text-driven 4D scene editing. Our method augments 3D Gaussians with quantized CLIP features to form a language-embedded dynamic representation, enabling efficient semantic querying of arbitrary spatial regions. We further propose a two-stage point-level localization strategy that first selects candidate Gaussians via CLIP similarity and then refines their spatial extent to improve accuracy. Finally, targeted edits are performed on localized regions using a diffusion-based video editing model, with flow and scribble guidance ensuring spatial fidelity and temporal coherence. 
Extensive experiments demonstrate that \sysname{} enables high-quality, text-driven edits across diverse scenes and object types, while preserving the appearance and geometry of unedited areas and surpassing prior approaches in both flexibility and visual fidelity.
\end{abstract}

\section{Introduction}

\label{sec:intro}
Neural Radiance Fields (NeRF)~\citep{mildenhall2021nerf} and 3D Gaussian Splatting (3DGS)~\citep{kerbl20233d} have emerged as powerful representations for modeling photorealistic 3D scenes, which have enabled a wide range of downstream applications, including novel view synthesis, relighting, semantic embedding, and scene editing. In particular, 3DGS achieves high-fidelity real-time rendering and has quickly become a practical tool for 3D content creation.
To reflect the dynamic nature of the real world, recent works have extended these static representations to dynamic scenes~\citep{fang2022fast, yang2024deformable, stearns2024dynamic}, supporting tasks such as dynamic reconstruction, motion tracking, and animation. In this work, we specifically focus on 4D scenes reconstructed from monocular videos, which are more practical in real-world scenarios but inherently more challenging than multi-view settings due to limited observations and ambiguities in geometry and motion.

Despite recent progress, editing 4D scenes remains a challenging and underexplored problem. 
A key limitation is the lack of fine-grained control over arbitrary objects or regions specified by the natural language. 
Recent works~\citep{mou2024instruct, he2024ctrl} enable text-driven editing in 4D scenes, 
but rely on 2D diffusion models to guide edits, which often introduce unintended modifications in irrelevant regions due to the absence of precise localization mechanisms.
In contrast, latest methods for static scenes~\citep{xu2024tiger, zhang20243ditscene, zhuang2024tip, liu2024sketchdream} demonstrate that text-driven editing of specific objects or parts is feasible when accurate spatial localization is available. 
However, these techniques are not designed for monocular dynamic settings and fail to account for object motion or temporal consistency. 
Ideally, a 4D editing system should allow users to describe desired changes with text prompts, selectively modify only relevant regions in space and time, and preserve the appearance and motion of all other content.

To address these challenges, we propose \sysname{}, a text-driven editing framework for 4D scenes reconstructed from monocular videos. 
Our key idea is to represent dynamic scenes with \textit{language-embedded Gaussians}, where each 3D Gaussian is enriched with compact CLIP-based semantic features. This representation enables semantic localization directly in 3D, forming the basis for point-level localization and text-driven editing. 
In addition, diffusion-based video editing models provide richer control conditions and can generate high-quality, temporally consistent edits from text prompts, making them particularly beneficial for dynamic scene editing and well-suited to our monocular setting.
Specifically, our approach consists of three main stages: (1) we construct a dynamic 3D Gaussian field augmented with quantized CLIP features, enabling each Gaussian to carry natural language semantics; (2) we introduce a point-level localization module that identifies and refines Gaussians relevant to a user-provided text query by combining 2D semantic supervision with 3D decoding; (3) we update only the localized Gaussians using guidance from a diffusion-based video editing model, leveraging optical flow and scribble to maintain spatial precision and temporal coherence. This pipeline allows for high-fidelity, region-specific edits while preserving the motion and appearance of unedited regions.

We evaluate \sysname{} on a diverse set of dynamic scenes that encompass both foreground and background elements with varying appearances and motion patterns. As shown in the experiments, our method accurately localizes target regions and generates high-quality, temporally coherent 4D edits while preserving the background, highlighting the flexibility and effectiveness of our framework for text-driven 4D scene editing.

This work specifically focuses on editing 4D scenes reconstructed from monocular videos, a practical yet challenging setting that differs from multi-view scenarios by providing only limited observations. 
Our contributions are threefold:
\begin{itemize} [leftmargin=*]
\item First, we propose a unified framework for text-driven 4D scene editing that integrates language-embedded Gaussian representations with diffusion-based video editing models, enabling precise and temporally consistent edits. 
\item Second, we develop a novel point-level localization strategy as a key component of this framework, which accurately identifies and refines editable regions to support flexible semantic control. 
\item Third, we conduct comprehensive experiments and ablation studies, showing that combining language-embedded Gaussians with video editing models enables region-specific, temporally coherent edits while preserving unedited content.
\end{itemize}

\section{Related Work}
\label{sec:rw}

\subsection{Radiance Fields for Dynamic Scenes}
Given the success of NeRF, several works have extended it by introducing spatiotemporal fields~\citep{fridovich2023k, shao2023tensor4d} or deformation fields~\citep{park2021hypernerf, pumarola2021d, fang2022fast} to model scene dynamics.
Recent works based on 3DGS have achieved more efficient reconstruction and rendering of dynamic scenes by using deformation fields~\citep{yang2024deformable, huang2024sc} or high-dimensional Gaussian fields~\citep{duan20244d}. Meanwhile, many works ~\citep{som2024, lei2024mosca, stearns2024dynamic} have integrated priors such as camera estimation, depth estimation, and 2D tracking for initialization, and have learned a dynamic Gaussian field, realizing dynamic scene reconstruction from monocular video.

\subsection{Language-Embedded Radiance Fields}
Recent work has explored embedding language features into neural scene representations to enable semantic querying and editing. LERF~\citep{kobayashi2022decomposing, kerr2023lerf} injects CLIP features into NeRF for 3D region retrieval and interaction. Later works extend this idea to 3D Gaussians~\citep{shi2024language, qin2024langsplat, jiang2023feature}, achieving faster and more accurate querying by attaching language features to Gaussians.
However, these methods decode language queries in 2D space and cannot directly identify relevant Gaussians in 3D. OpenGaussian~\citep{wu2024opengaussian} introduces point-level localization but is limited to static scenes. 4D-LangSplat~\citep{li20254dlangsplat4dlanguage} and Feature4X~\citep{zhou2025feature4x} embeds semantics in dynamic Gaussians but lacks point-level precision and is not suitable for 4D scenes editing.
In contrast, our method achieves accurate point-level localization of language-embedded Gaussians, enabling precise 4D editing.

\subsection{Text-driven Radiance Fields Editing}
Text-driven editing has been explored in static 3D scenes using image diffusion models~\citep{haque2023instruct, chen2024gaussianeditor}, where edits are guided by text but lack spatial precision, resulting in unexpected changes outside the target area. In addition, recent methods for static scenes~\citep{zhuang2023dreameditor, xu2024tiger, zhang20243ditscene, zhuang2024tip, liu2024sketchdream} demonstrate that fine-grained, text-driven editing is achievable with accurate semantic grounding.
To handle dynamic content, recent works~\citep{he2024ctrl, mou2024instruct, shao2024control4d, kwon2025instruct} extend editing to 4D by combining frame-wise diffusion with temporal constraints, which can't accurately localize editing regions in 3D.
Instruct-4DGS~\citep{kwon2025instruct} further needs multi-view videos as input.
Building on this insight, our method introduces point-level localization of language-embedded Gaussians and leverages the video diffusion model, enabling selective, text-guided edits with high spatial precision and temporal coherence in 4D scenes.
The related works on text-driven video editing are discussed in Appendix~\ref{sub_sec:more_related}.

\section{Method}
\label{sec:method}

The input to our method is a monocular video of a 4D scene, along with a natural language prompt describing the desired edit. 
Our goal is to modify any user-specified region of the scene based on a text prompt, while leaving the rest of the scene untouched.
The target region can belong to any part of the scene, including static or dynamic elements.

\subsection{Preliminary: 3D Gaussian Splatting}
\label{sec:preli}

3D Gaussian Splatting~\citep{kerbl20233d} synthesizes photorealistic scenes by aggregating numerous colored 3D Gaussians, which are projected onto image planes via differentiable rasterization. Specifically, a 3D scene is represented by a set of Gaussians $\mathcal{G}$, where each Gaussian is parameterized by its center $\boldsymbol{\mu} \in \mathbb{R}^3$, rotation $\mathbf{R} \in \mathrm{SO}(3)$, scale $\boldsymbol{s} \in \mathbb{R}^3$, opacity $o \in \mathbb{R}$, and color $\boldsymbol{c} \in \mathbb{R}^3$. 
Given a camera $\mathcal{C}$ with known intrinsics and extrinsics, the Gaussians are then projected onto the image plane and composited through a differentiable rasterizer $\mathcal{R}$, yielding the final image:
\begin{equation} 
I = \mathcal{R}(\boldsymbol{\mu}, \mathbf{R}, \boldsymbol{s}, o, \boldsymbol{c}; \mathcal{C}) . \label{eq:render}
\end{equation}

\begin{figure*}
  \centering
  \includegraphics[width=\textwidth]{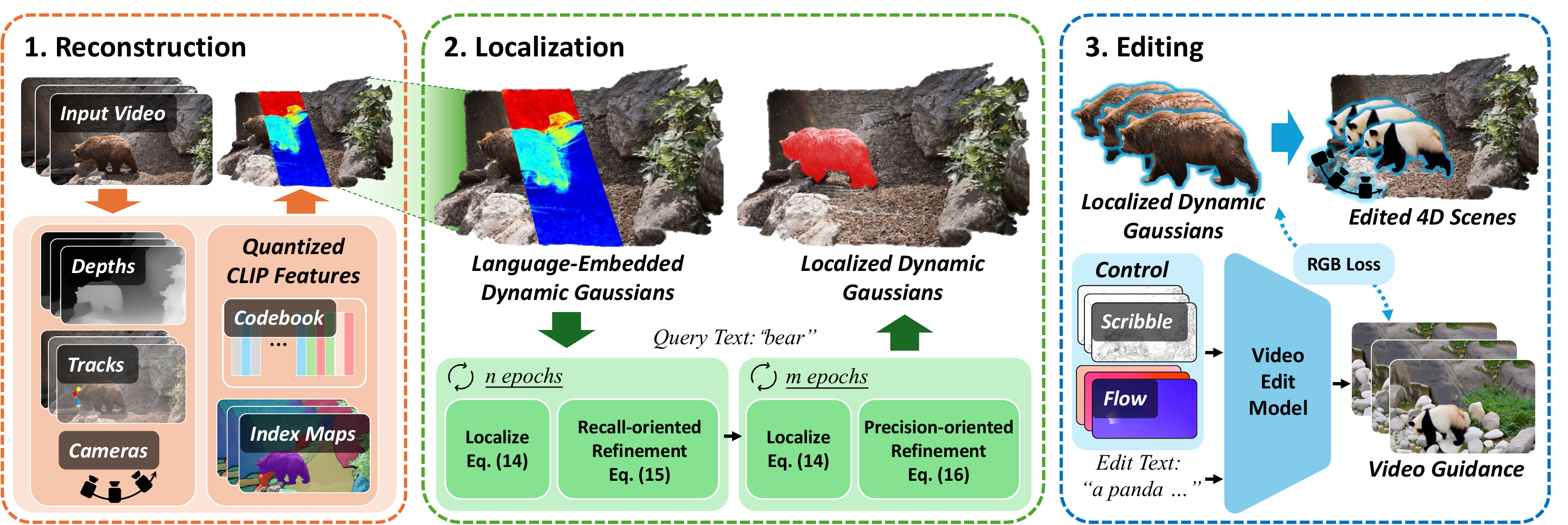}
  \caption{Overview of our method. Given a monocular video 
  , we construct a Language-Embedded Dynamic Gaussian field by enriching 3D Gaussians with quantized CLIP features (Section~\ref{sec:ledg}). We then perform point-level localization to identify Gaussians relevant to the query, using 2D relevance maps and 3D semantic decoding (Section~\ref{sec:localization}). Finally, we apply text-driven editing with a diffusion-based video model, modifying only the localized Gaussians to produce temporally consistent and spatially precise edits (Section~\ref{sec:editing}). 
  The colored visualization in the Language-Embedded Dynamic Gaussian field shows PCA results after semantic feature splatting.
  }
  \label{fig:pipeline}
\end{figure*}

\subsection{Language-Embedded Dynamic Gaussians}
\label{sec:ledg}

Inspired by recent advances in language-embedded fields~\citep{kerr2023lerf, shi2024language, qin2024langsplat}, we propose a 4D Gaussian field enriched with semantic features from the input monocular video. 
By embedding language semantics into dynamic Gaussians, we enable region-specific editing through natural text guides. This approach allows for precise modifications to targeted areas, while preserving the unchanged content by freezing non-target Gaussians.

Previous work~\citep{zhou2025feature4x} embeds language features into dynamic Gaussians from monocular video, enabling tasks such as segmentation. However, this approach often sacrifices rendering efficiency and semantic fidelity due to the complex feature distillation and interpolation process. 
In contrast, we adopt a quantization-based feature compression strategy from LEGaussians~\citep{shi2024language}, which efficiently encodes CLIP features while maintaining high-fidelity reconstruction and enabling faster training, making it more suitable for real-time text-driven dynamic scene editing.

\noindent \textbf{Data preprocessing.}
To reconstruct the dynamic field with semantic features from the input video, we apply standard preprocessing steps: (1) extracting the camera intrinsics and extrinsics for each frame, (2) extracting dynamic masks, monocular depth, and long-range 2D point tracks for foreground pixels in each frame, and (3) obtaining pixel-level CLIP features for each frame.
The first two pre-processing steps are crucial for reconstructing the dynamic Gaussian field from monocular video. We refer to previous works~\citep{som2024, stearns2024dynamic, lei2024mosca} to extract camera poses and obtain dynamic masks, depth, and 2D tracks using off-the-shelf methods.

Step three serves as the foundation for embedding semantics into the dynamic Gaussian field. First, we use SAM2~\citep{ravi2024sam} to generate multi-class tracking masks for the video. 
For each frame, we crop the image using these masks and then pass the cropped regions through the CLIP image encoder~\citep{radford2021learning} to obtain feature embeddings. These CLIP features are assigned to the corresponding pixels in the mask, embedding semantic information at the pixel level. 
The video consists of $t$ frames, each of size $h \times w$, and the CLIP features for each frame are extracted at the pixel level, resulting in a feature tensor $F_{\text{clip}} \in \mathbb{R}^{t \times h \times w \times c}$, where $c$ represents the number of CLIP feature channels.
Following the approach in LEGaussians~\citep{shi2024language}, we quantize the extracted CLIP features using a learnable codebook $\mathcal{B} \in \mathbb{R}^{N \times c}$, where $N$ is the number of codebook entries.
After quantization, we obtain a learned codebook and the corresponding index map $M_{\text{index}} \in \mathbb{R}^{t \times h \times w \times 1}$, which stores the closest codeword index for each pixel in each frame.
These index maps and the video-level semantic codebook $\mathcal{B}$ are used to embed semantics into the dynamic Gaussian field. More details of the quantization of CLIP features will be presented in Appendix~\ref{sec_app:quan_detail}. 

\noindent \textbf{Language-Embedded Dynamic Gaussians.}
Following the Shape-of-Motion~\citep{som2024}, we represent motion as a rigid transformation in $\mathrm{SE}(3)$ applied to canonical 3D Gaussians to model dynamic parts of scenes, and the others remain static. At time $t$, the transformed position and orientation are 
$\boldsymbol{\mu}_t = \mathbf{R}_t \boldsymbol{\mu} + \mathbf{t}_t$
and 
$\mathbf{R}_t = \mathbf{R}_t \mathbf{R}$,
where $\mathbf{T}_t = (\mathbf{R}_t, \mathbf{t}_t) \in \mathrm{SE}(3)$ denotes the rigid transformation from the canonical frame to time $t$. We set the first frame to the canonical frame. To regularize motion and reduce overfitting, a low-dimensional parameterization is adopted by introducing $B$ global motion bases $\{ \mathbf{T}^{(b)}_t \}_{b=1}^B$ shared across all dynamic Gaussians. The transformation at time $t$ is then expressed as a weighted sum:
$\mathbf{T}_t = \sum_{b=1}^B w^{(b)} \mathbf{T}^{(b)}_t, \text{with} \sum_{b=1}^B w^{(b)} = 1$, where $w^{(b)}$ are per-Gaussian, learnable coefficients.

We assign each 3D Gaussian a learnable semantic feature vector $\boldsymbol{f} \in \mathbb{R}^{d_f}$ to encode compact language semantics. Directly using discrete indices $M_{\text{index}}$ is not compatible with differentiable rendering, so we instead render these continuous features and supervise them using the quantized semantic map.
At time $t$, each dynamic Gaussian has transformed parameters $\boldsymbol{\mu}_t$ and $\mathbf{R}_t$. These are used to render a 2D semantic feature map $I_f^{(t)}$ through differentiable rasterization $\mathcal{R}$, analogously to the photometric rendering in Eq.~\ref{eq:render}:
\begin{equation}
I_f^{(t)} = \mathcal{R}(\boldsymbol{\mu}_t, \mathbf{R}_t, \boldsymbol{s}, o, \boldsymbol{f}; \mathcal{C}_t),
\label{eq:render_lang}
\end{equation}
where $\boldsymbol{f}$ replaces color as the per-Gaussian channel to be composited, and $\mathcal{C}_t$ denotes the camera at frame $t$. The rendered feature map $I_f^{(t)} \in \mathbb{R}^{H \times W \times d_f}$ is then decoded into a semantic index distribution via a lightweight MLP $\mathcal{D}$:
\begin{equation}
\hat{M}^{(t)} = \text{softmax}(\mathcal{D}(I_f^{(t)})) \in \mathbb{R}^{H \times W \times N},
\label{eq:decode_lang}
\end{equation}
where $N$ is the number of codebook entries in $\mathcal{B}$, and $\hat{M}^{(t)}$ represents the predicted distribution over discrete semantic indices.

\noindent \textbf{Language Embedding Loss.}
To supervise the learning of semantic features, we use a cross-entropy loss between the predicted semantic index distribution $\hat{M}^{(t)}$ and the ground-truth index map $M_{\text{index}}^{(t)}$ obtained during feature quantization:
\begin{equation}
\mathcal{L}_{\text{lang}} = \text{CE}(\hat{M}^{(t)}, M_{\text{index}}^{(t)}).
\label{eq:lang_loss}
\end{equation}
This objective encourages each dynamic Gaussian to encode a compact, differentiable semantic descriptor that aligns with language embeddings, enabling region-specific manipulation via natural language commands.

\noindent \textbf{Reconstruction Loss.}
The reconstruction loss $\mathcal{L}_{\text{rec}}$ consists of four components: RGB loss, depth loss, mask loss, and tracking loss. These losses supervise the appearance, geometry, and motion of the scene and follow the implementation in Shape-of-Motion~\citep{som2024}. 

\noindent \textbf{Optimization.}
We optimize the Language-Embedded Dynamic Gaussians by minimizing a weighted combination of the reconstruction loss and the language embedding loss:
\begin{equation}
\mathcal{L} = \lambda_{\text{rec}} \mathcal{L}_{\text{rec}} + \lambda_{\text{lang}} \mathcal{L}_{\text{lang}},
\label{eq:ledg_loss}
\end{equation}
where $\lambda_{\text{rec}}$ and $\lambda_{\text{lang}}$ are weights that balance the two objectives.

\subsection{Point-level Localization of Gaussians}
\label{sec:localization}
While Language-Embedded Dynamic Gaussians encode semantic features on each 3D Gaussian, these features only acquire meaning after being splatted onto the 2D image plane. In isolation, the per-Gaussian embeddings lack explicit correspondence to the semantics of real-world objects. Consequently, directly localizing and editing Gaussians based on text remains inaccurate and coarse.

Existing methods such as OpenGaussian~\citep{wu2024opengaussian} optimize the entire scene representation to align with semantics, which is inefficient and lacks granularity for object-specific editing. In contrast, we introduce a point-level localization framework that accurately localize Gaussians relevant to the user query, by combining 2D semantic supervision and 3D feature decoding. We further refine the localization through a two-stage optimization that improves both recall and precision. This approach enables efficient and accurate text-driven editing by isolating only the relevant Gaussians without affecting unrelated regions.

Given a user-provided query text $q$, our goal is to localize all Gaussians that correspond to the described object at a fine-grained point-level resolution. We achieve this by leveraging the semantic features learned in Section~\ref{sec:ledg}.

\noindent \textbf{Language-guided Localization in 2D and 3D.}
At each frame $t$, we render a 2D semantic feature map $I_f^{(t)} \in \mathbb{R}^{H \times W \times d_f}$ from the trained Language-Embedded Dynamic Gaussians and decode it into CLIP space using the trained decoder and codebook $\mathcal{B}$ (as in Eq.~\ref{eq:render_lang} and Eq.~\ref{eq:decode_lang}).
Then, we compute the relevance map between the image-aligned CLIP feature and the query text feature $F_q$ using cosine similarity, following the approach of Kerr et al.~\citep{kerr2023lerf}:
\begin{equation}
R^{(t)}(p) = \cos\left(\hat{M}^{(t)}(p) \cdot \mathcal{B}, F_q\right),
\label{eq:relevance_map}
\end{equation}
where $\hat{M}^{(t)}$ is the predicted index distribution over codebook entries, and $p$ is a 2D pixel location.

We threshold this relevance map with a hyperparameter $\tau$ to obtain a binary 2D mask $M_{\text{2D}}^{(t)}$ indicating regions related to the text:
\begin{equation}
M_{\text{2D}}^{(t)}(p) = \mathbf{1}\left[ R^{(t)}(p) > \tau \right], 
\label{eq:2d_mask}
\end{equation}
where $\mathbf{1}[\cdot]$ is the indicator function.

We also perform text-based localization directly in 3D. For each Gaussian $g_i$ with semantic feature $\boldsymbol{f}_i$, we first obtain the codebook index distribution $\hat{m}_i$ as in Eq.~\ref{eq:decode_lang} using the decoder, and decode it with the codebook to get its CLIP-space embedding $\tilde{f}_i = \mathcal{B} \cdot \hat{m}_i \in \mathbb{R}^{c}$.
Then, we compute its cosine similarity with the query:
\begin{equation}
r_i = \cos\left(\tilde{f}_i, F_q\right).
\label{eq:similarity_3d}
\end{equation}

We define a 3D Gaussian mask $\mathbf{L}_{\text{3D}} = \{ g_i \mid r_i > \tau \}$, indicating Gaussians localized by the query text. We denote the entire 3D localization pipeline as a function:
\begin{equation}
\mathbf{L}_{\text{3D}} = \texttt{Localize}(q, \tau),
\label{eq:localize_fn}
\end{equation}
which returns a set of Gaussians matching the query.

\noindent \textbf{Recall-oriented Refinement.}
The initial $\texttt{Localize}(q, \tau)$ function may miss relevant Gaussians (false negatives). To recover these, we render the complement set $\bar{\mathbf{L}}_{\text{3D}} = \{ g_i \mid r_i \leq \tau \}$ and project their features to 2D. We then restrict the optimization to pixels inside the 2D relevance mask $M_{\text{2D}}^{(t)}$:
\begin{equation}
\mathcal{L}_{\text{recall}} = \text{CE}(\mathcal{D}(I_f^{(t)}), M_{\text{index}}^{(t)}) \quad \text{for } p \in M_{\text{2D}}^{(t)}.
\label{eq:loss_recall}
\end{equation}
This loss encourages previously missed Gaussians to move closer to the query’s semantic space. We repeat this process for $n$ epochs to improve recall.

\noindent \textbf{Precision-oriented Refinement.}
While the recall stage recovers most relevant Gaussians, it may also introduce unrelated ones (false positives). To refine precision, we freeze the correctly localized Gaussians inside the 2D mask and optimize only those outside:
\begin{equation}
\mathcal{L}_{\text{precision}} = \text{CE}(\mathcal{D}(I_f^{(t)}), M_{\text{index}}^{(t)}) \quad \text{for } p \notin M_{\text{2D}}^{(t)}.
\label{eq:loss_precision}
\end{equation}
This stage suppresses irrelevant Gaussians by aligning their features back to their original semantics. The optimization runs for $m$ epochs.

\noindent \textbf{Final Localization.}
After alternating between recall and precision stages, we apply the $\texttt{Localize}(q, \tau)$ function again to obtain the final point-level Gaussian mask corresponding to the user query. This localization process is more accurate and semantically coherent, enabling fine-grained region editing with natural language.

\subsection{Text-driven Editing with Video Model}
\label{sec:editing}

Recent video editing models~\citep{jiang2025vace, bian2025videopainter} offer a variety of control modalities, including masks, optical flow, scribble, and grayscale frames, to guide generative edits. Among them, mask-based control allows restricting edits to spatial regions, but lacks the ability to influence motion and appearance details within the masked area. Conversely, optical flow and scribble offer richer motion and structure cues, leading to more coherent and realistic edits, but these methods affect the entire video and cannot restrict edits to only a specific region.

Our approach aims to combine the spatial precision of mask-based editing with the rich motion and appearance details offered by flow- and scribble-based controls. By utilizing 3D Gaussians localized in Section~\ref{sec:localization}, we can selectively optimize the desired parts of the scene while preserving the overall structure and motion. This strategy allows us to benefit from high-fidelity editing signals without compromising spatial specificity, enabling localized, realistic edits in dynamic scenes.

\noindent \textbf{Video Guidance for Gaussians Editing.}
We adopt the diffusion-based video editing model VACE~\citep{jiang2025vace}, which supports text-driven editing guided by auxiliary signals. To strike a balance between control strength and generative flexibility, we primarily use optical flow and scribble, which guide edits while preserving the underlying motion and scene structure.
The control signals are extracted from the original input video. Optical flow is computed using RAFT~\citep{teed2020raft} and scribble are extracted using ~\citep{chan2022learning}.
These control conditions, along with the original video and text prompt $q$, are used to generate an edited video $V_{\text{edit}}$, which serves as a reference for editing the localized Gaussians.

\noindent \textbf{Localized Gaussian Editing.}
Given the localized set $\mathbf{L}_{\text{3D}}$ from Eq.~\ref{eq:localize_fn}, we freeze all Gaussians outside $\mathbf{L}_{\text{3D}}$ and only update those within the set. This ensures that editing is confined to semantically relevant content, while preserving the rest of the scene.

\noindent \textbf{Optimization Procedure.}
The process described here is part of training, where we optimize only the Gaussians within $\mathbf{L}_{\text{3D}}$. We use the video $V_{\text{edit}}$, as a reference for updating the Gaussians. Specifically, we render a video $V_{\text{render}}$ from the dynamic 3D Gaussians and minimize the pixel-wise loss between the rendered video and the edited video:
\begin{equation}
\mathcal{L}_{\text{edit}} = \| V_{\text{render}} - V_{\text{edit}} \|_2^2.
\label{eq:edit_loss}
\end{equation}
During this process, the gradients are only propagated to the Gaussians in $\mathbf{L}_{\text{3D}}$, ensuring that the editing is focused on the selected regions. This optimization typically proceeds for $k$ epochs. After the optimization, the dynamic Gaussians can be used for rendering, producing the final edited result.

\section{Experiments}
\label{sec:exper}

\subsection{Experimental Setup}

\label{sub_sec:experiment}
\noindent \textbf{Dataset.}
To comprehensively assess the effectiveness of our method, we organize experiments on scenes from DAVIS~\citep{Caelles_arXiv_2019}, DyCheck iPhone~\citep{gao2022dynamic} datasets, Dynerf datasets ~\citep{li2022neural} and some videos collected from wild. These scenes, captured by a monocular camera, encompass a rich diversity of objects, 
set against varying backgrounds. We utilize the DyCheck iPhone datasets for comparison with baselines and employ the other datasets to demonstrate the feasibility of our method on casual monocular videos.

\noindent \textbf{Baselines.}
We compare our method against two recent approaches for text-driven dynamic scene editing. 
(1) \textbf{Instruct 4D-to-4D (IN4D)}~\citep{mou2024instruct} enhances InstructPix2Pix~\citep{brooks2023instructpix2pix} with anchor-aware attention and optical flow-guided propagation to achieve temporally consistent video edits, treating 4D scenes as pseudo-3D volumes. 
(2) \textbf{CTRL-D}~\citep{he2024ctrl} performs the editing by fine-tuning InstructPix2Pix on a reference image and then optimizing deformable 3D Gaussians in two stages, enabling controllable and consistent 4D scenes editing.

\noindent \textbf{Metrics.}
Following prior works~\citep{haque2023instruct, zhuang2023dreameditor}, we assess editing quality using CLIP Text-Image directional similarity~\citep{gal2022stylegan}, which quantifies the consistency between the intended textual edit and the resulting visual change. 
For fairness, all methods generate output videos under a shared camera trajectory. 
The directional similarity is computed per frame and averaged across time to obtain the final score.
To complement the automatic metric with perceptual insights,  we conduct a user study. Participants view edited dynamic scenes from all methods under a rotating viewpoint and select the most satisfying result. We collect 58 responses and report the voting ratio per method.
Quantitative comparisons use 3 scenes covering 9 distinct editing operations.
Implementation details are included in Appendix~\ref{sec_app:imple_detail}.

\begin{figure}[t]
  \centering
  \includegraphics[width=\textwidth]{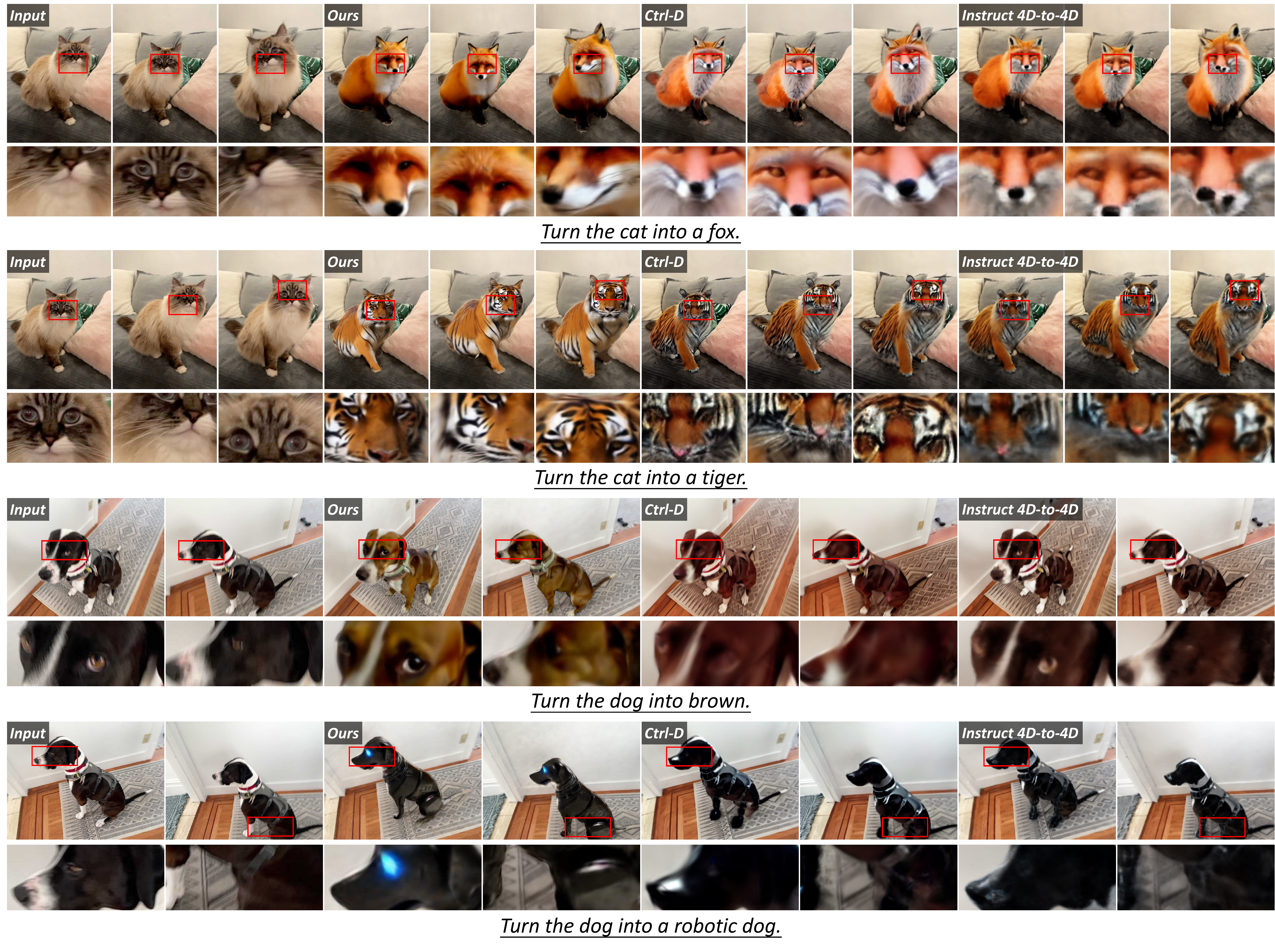}
  \caption{
    Comparison of editing results on the iPhone dataset.
    Our method achieves better temporal coherence, finer details (e.g., whiskers, eyes, specular highlights), and more accurate motion, while avoiding artifacts in unrelated regions.
    Baselines tend to over-edit or introduce distortions due to limitations in 2D diffusion-based approaches.
  }
  \label{fig:comparison}
  \vspace{-5mm}
\end{figure}

\subsection{Qualitative Results}

\noindent \textbf{Editing 4D Scenes.}
We present qualitative results of text-driven editing in 4D scenes from both DAVIS and iPhone videos, as shown in Figure~\ref{fig:teaser}. The figure illustrates the input monocular video and the output edited scene rendered from multiple novel views in two different time steps.
Thanks to our point-level localization mechanism, \sysname{} is able to precisely identify and edit only the target regions described by the text prompt, while preserving all unrelated content. Furthermore, by incorporating a video editing model guided by optical flow and scribble, our method produces realistic and temporally coherent modifications. The resulting edits not only match the spatial semantics but also maintain consistent motion across time and viewpoints, demonstrating the robustness of our text-driven 4D scene editing pipeline. More results are in Appendix~\ref{sec_app:more_editing_results} and Appendix~\ref{sec_app:semantic_query_localization}.

\noindent \textbf{Comparison with Baseline Methods.}
We compare \sysname{} with two baselines, IN4D and CTRL-D, on the iPhone dataset (Figure~\ref{fig:comparison}). We evaluate two scenes: one involving a cat and another featuring a dog, each with two distinct editing prompts. For the cat scene, each method's result is shown in three time steps; for the dog scene, in two time steps.
\sysname{} consistently outperforms the baselines in both temporal consistency and spatial accuracy. For example, our method preserves sharp, temporally aligned textures when editing a cat into a tiger, while baselines show flickering and misalignments despite handcrafted temporal modules. Additionally, \sysname{} produces finer details, such as the whiskers on the edited fox or the glossy highlights on the robotic dog's leg, while maintaining realistic appearance across time. 
Our motion modeling is also more accurate, as shown in the cat scene where baselines bias the head orientation due to 2D editing limitations.
Crucially, \sysname{} does not introduce artifacts in unrelated regions of the scene. In contrast, baselines often modify background elements (e.g. the ground) when editing an object, such as editing a dog into a robotic version, leading to noticeable texture inconsistencies. Moreover, baseline approaches can inadvertently affect the overall color scheme of the scene when modifying the color of specific objects, such as in the "Turn the dog brown" task.

\subsection{Quantitative Results}

\begin{wraptable}{r}{0.45\textwidth} 
  \centering
  \vspace{-20pt} 
  \caption{
    Quantitative comparison of editing methods based on CLIP similarity and user preference. The CLIP similarity is computed as the average text-image similarities between the prompt and all output video frames.
  }
  \vspace{5pt}
  \label{tab:quantitative}
  \resizebox{0.38\textwidth}{!}{
      \begin{tabular}{lcc} 
        \toprule
        Method & CLIP Sim. $\uparrow$ & User Pref. $\uparrow$ \\
        \midrule
        IN4D   & $25.24$ & $28.62\%$ \\
        CTRL-D & $26.04$ & $28.39\%$ \\
        Ours   & $\mathbf{26.23}$ & $\mathbf{42.99\%}$ \\
        \bottomrule
      \end{tabular}
  }
\end{wraptable}

Table~\ref{tab:quantitative} presents the quantitative results in the iPhone dataset between our method and two baselines. We measure the CLIP similarity between the editing results and the intended text prompts. The results indicate that our method outperforms existing approaches in terms of alignment between the results and the edited prompts. Additionally, a crowd-sourced subjective evaluation is conducted, with further details provided in Appendix~\ref{sec_app:detailed_quan_res}. The user preference scores demonstrate our method achieves superior visual perceptual quality.

\subsection{Ablation Study}

We perform an ablation study to assess the contribution of the two localization refinement steps in our point-level localization module: \textit{R-Ref} (Recall-oriented Refinement) and \textit{P-Ref} (Precision-oriented Refinement). We compare four variants: the full pipeline (\textit{Full}), and three simplified versions: (\textit{w/o Ref}), where both refinement are removed; (\textit{w/o R-Ref}); and (\textit{w/o P-Ref}). In the \textit{w/o Ref} setting, we apply the localization function defined in Eq.~\ref{eq:localize_fn} to directly select Gaussians.

\begin{figure*}[t]
  \centering
  \includegraphics[width=0.9\textwidth]{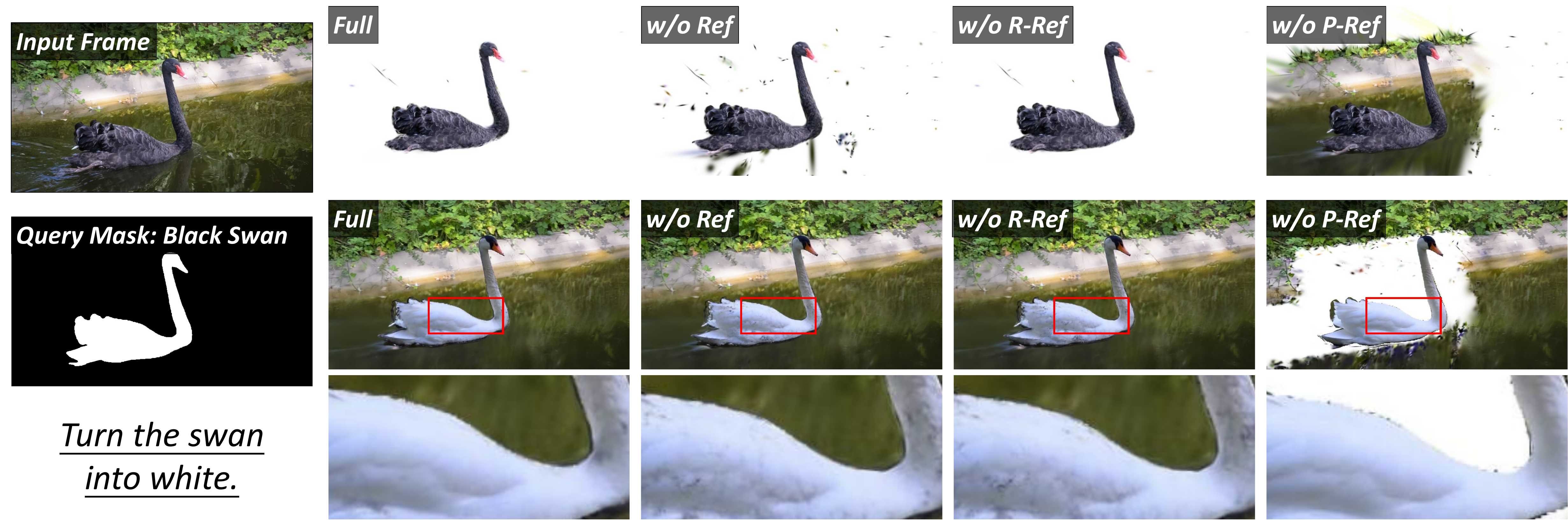}
  \vspace{-3mm}
  \caption{
    Qualitative ablation study on the effect of localization refinement, including (1) input frame, (2) query mask based on the input prompt, (3) localized Gaussian rendering and edited result of the \textit{Full} model, (4) result of \textit{w/o Ref}, (5) result of \textit{w/o R-Ref} and (6) result of \textit{w/o P-Ref}.
  }
  \vspace{-3mm}
  \label{fig:ablation}
\end{figure*}

Figure~\ref{fig:ablation} shows qualitative comparisons of a DAVIS dataset dynamic scene, visualizing localized regions and edit results for each method.  Compared to variants, the \textit{full} method confines edits more accurately to the intended region and better preserves the appearance of unrelated areas. 
Omitting the refinement steps (\textit{w/o Ref}) leads to over-selection of irrelevant Gaussians, causing unintended changes and degraded rendering quality. Removing only R-Ref (\textit{w/o R-Ref}) leads to incomplete Gaussian localization, for example, residual dark spots remain on the body of the swan. Conversely, removing \textit{P-Ref} (\textit{w/o P-Ref})  broadens selection coverage, but introduces noisy, unrelated Gaussians. This causes large portions of the background to be incorrectly included in the edit region, leading to visible blank artifacts in the final result. These observations highlight the complementary roles of \textit{R-Ref} and \textit{P-Ref} in ensuring accurate and clean localization.

\begin{wraptable}{r}{0.4\textwidth} 
  \centering
  \vspace{-20pt} 
  \caption{
    Ablation studies on the DAVIS dataset.
  }
  \vspace{5pt}
  \label{tab:ablation}
  \resizebox{0.38\textwidth}{!}{
      \begin{tabular}{lccc}
        \toprule
        Variant & PSNR $\uparrow$ & mIoU (\%) $\uparrow$ \\
        \midrule
        w/o R-Ref & 39.43 & \textbf{0.71} \\
        w/o P-Ref  & 29.70 & 0.13 \\
        w/o Ref & 36.86 & 0.43 \\
        Full (Ours) & \textbf{39.55} & \textbf{0.71} \\
        \bottomrule
      \end{tabular}
    }
\end{wraptable}

We further quantify this behavior in Table~\ref{tab:ablation}, using two metrics: (1) PSNR between the localized Gaussian rendering and the reconstruction within the query mask, and (2) mIoU between the localized Gaussians and the text queried mask. Our method achieves superior performance on all metrics, demonstrating the importance of refinement in achieving accurate localization of Gaussians based on query text, which benefits localized edits and preserves background fidelity.

The ablation studies of video editing models are presented in Appendix~\ref{sec_app:ablation_of_video}.

\section{Conclusion and Limitation}
\label{sec:conclu}

\sysname{} demonstrates that enriching dynamic 3D Gaussians with language-aligned features enables flexible and precise text-driven editing of 4D scenes reconstructed from monocular videos. By integrating a language-embedded Gaussian representation, point-level localization, and diffusion-based video editing, our framework achieves high-quality, temporally coherent, and region-specific edits while preserving unedited content. 
This shows that semantics can be embedded into dynamic scene representations, opening new opportunities for controllable and interactive 4D content creation. 
Despite these advances, our method is still constrained by the limitations of monocular reconstruction: the underlying Gaussian representation we adopt only supports monocular dynamic scene reconstruction, which restricts our framework to monocular videos. In addition, reconstruction may suffer from depth and pose errors in highly dynamic scenarios, and diffusion-based models can introduce temporal drift or motion artifacts. 
Future work will focus on extending our approach to multi-view 4D representations, improving monocular reconstruction under challenging motion and enhancing generative models with stronger motion fidelity and regional control.

\bibliography{ref}
\bibliographystyle{ref}

\clearpage
\appendix
\section{More Related Works}
\label{sub_sec:more_related}

\subsection{Text-driven Video Editing}
Video editing models are the foundation of our \sysname{}. Several generative video editing methods utilize training-free adaptations of pre-trained Text-to-Image models, which typically convert spatial self-attention mechanisms into temporal-aware cross-frame attention to enforce consistency~\citep{ceylan2023pix2video, khachatryan2023text2video, qi2023fatezero, geyer2023tokenflow, yang2023rerender}. Another strategy is per-video fine-tuning, where a image generation model is optimized on a specific input video to learn its unique appearance and motion dynamics~\citep{liu2024video, wu2023tune, shin2024edit}. More advanced methods like VACE~\citep{jiang2025vace}, by handling various multimodal inputs through a unified framework,  achieve versatile compositional editing. 
In our framework, we leverage the compositional and multimodal control capabilities of VACE to guide localized Gaussian updates, enabling spatially precise and temporally coherent text-driven 4D editing.

\section{Technical Details}

\subsection{Quantization of CLIP Features}
\label{sec_app:quan_detail}
Following the approach in LEGaussians~\citep{shi2024language}, we quantize the extracted CLIP features $F_{\text{clip}} \in \mathbb{R}^{t \times h \times w \times c}$ using a learnable codebook $\mathcal{B} \in \mathbb{R}^{N \times c}$, where $N$ is the number of codebook entries.
For each frame $i$ and pixel location $p$, we compute the cosine similarity between the CLIP feature vector $F_{\text{clip}}^{(i)}(p) \in \mathbb{R}^c$ and all entries in the codebook $\mathcal{B}$. The feature is then assigned the index of the most similar codebook entry:
\begin{equation}
M_{\text{index}}^{(i)}(p) = \arg\max_{j \in \{1, \dots, N\}} \, \cos\left(F_{\text{clip}}^{(i)}(p), \mathcal{B}^{(j)}\right),
\label{eq:quan}
\end{equation}
Here, $\mathcal{B}^{(j)} \in \mathbb{R}^c$ denotes the $j$-th codebook vector, and $M_{\text{index}}^{(i)}(p) \in \{1, \dots, N\}$ is the index of the closest codeword for pixel $p$ in frame $i$ based on cosine similarity. 

To optimize the codebook $\mathcal{B}$, we use a cosine similarity loss function, which encourages the codebook entries to move closer to the CLIP feature vectors:
\begin{equation}
\mathcal{L}_{\text{quant}} = \sum_{i=1}^{t} \sum_{p=1}^{h \times w} \left( 1 - \cos\left(F_{\text{clip}}^{(i)}(p), \mathcal{B}^{M_{\text{index}}^{(i)}(p)}\right) \right).
\end{equation}
This loss, inspired by the work of~\citep{van2017neural}, is backpropagated to update the codebook $\mathcal{B}$, allowing it to better represent the distribution of the extracted CLIP features. After training, we obtain a learned codebook and the corresponding index map $M_{\text{index}} \in \mathbb{R}^{t \times h \times w \times 1}$, which stores the closest codeword index for each pixel in each frame.

These index maps and the video-level semantic codebook $\mathcal{B}$ are used to embed semantics into the dynamic Gaussian field.

\subsection{Implementation Details}
\label{sec_app:imple_detail}
We build upon Shape-of-Motion~\citep{som2024}, originally built for monocular video, and extend it into a language-embedded dynamic 3D representation.
Given a monocular video, we extract camera poses via MegaSaM~\citep{li2024megasam} or Droid-slam~\citep{teed2021droid} and obtain dynamic masks, depth, and 2D tracks using off-the-shelf methods~\citep{ravi2024sam, li2024megasam, piccinelli2024unidepth, teed2020raft}.
Following~\citep{fan2024instantsplat}, we learn a camera pose correction term to refine the predicted poses, leading to improved dynamic scene reconstruction.
Semantic features are obtained from SAM2~\citep{ravi2024sam} and dense CLIP~\citep{radford2021learning}, then quantized with a codebook of size $N{=}128$~\citep{shi2024language}. 
In the phase of dynamic Gaussian reconstruction, we follow the parameter configurations specified in Shape-of-Motion~\citep{som2024}. We utilize the Adam Optimizer for the optimization process. Concretely, we carry out 1000 optimization iterations for the initial fitting procedure and 500 epochs for the joint optimization procedure. For the initialization of Gaussians, 40,000 dynamic Gaussians are initialized for the dynamic segment of the scene, while 100,000 static Gaussians are initialized for the static segment. Furthermore, we implement the identical adaptive Gaussian control for both dynamic and static Gaussians, in accordance with the method described in 3DGS~\citep{kerbl20233d}. Each 3D Gaussian is augmented with a learnable semantic vector ($d_f{=}8$). 
Training optimizes a weighted sum of $\mathcal{L}_{\text{rec}}$, $\mathcal{L}_{\text{lang}}$ (both weighted by 1), and geometry terms. 
Language-based selection uses similarity in the CLIP space, combining 2D and 3D signals with a threshold $\tau{=}0.95$. 
A two-stage refinement runs for $n{=}50$ epochs to improve recall, and additional denoising is performed for $m{=}10$ epochs. 
We train for $500$ epochs for language-embedded reconstruction and another $k{=}500$ epochs for editing on a single NVIDIA A6000.

\section{More Results}
\label{sub_app:more_results}

This section provides additional experimental results of our method. Appendix~\ref{sec_app:detailed_quan_res} presents a quantitative comparison of results between our method and the baselines; Appendix~\ref{sec_app:more_editing_results} demonstrates more editing outcomes of our approach across different datasets; Appendix~\ref{sec_app:semantic_query_localization} provides results of semantic querying and Gaussian point localization using text prompt; 
and Appendix~\ref{sec_app:ablation_of_video} reports the results of ablation experiments conducted on the video editing model.

\subsection{Detailed Quantitative Results}
\label{sec_app:detailed_quan_res}
To comprehensively evaluate the performance of our proposed method, we conducted a crowd-sourced subjective evaluation against the baseline approaches IN4D~\citep{mou2024instruct} and CTRL-D~\citep{he2024ctrl}. The study was designed to gather qualitative feedback across three distinct dimensions of video editing quality: Naturalness, which measures the realism of the edited results considering both spatial accuracy and temporal consistency; Prompt Fidelity, which assesses how well the edits match the given text instruction; and Background Preservation, which evaluates whether irrelevant regions remain unchanged.

For the evaluation, we curated nine different video editing scenarios. In each scenario, participants were presented with the results generated by the three methods (IN4D, CTRL-D, and Ours) in a randomized order. We then asked the participants to rank these three results from best (rank 1) to worst (rank 3) for each of the three evaluation dimensions. A total of 58 participants completed the survey for the three dimensions. Table~\ref{tab_app:user} presents a detailed breakdown of the results, showing the percentage of first-place votes each method received for each criterion. This quantitative data supplements the main findings in our paper, offering a more granular view of user preferences.
Results show that \sysname{} consistently outperforms the baselines across all three evaluation dimensions.

\begin{table}[htbp]
  \centering
  \caption{
    Results of the crowd-sourced subjective evaluation. The values indicate the percentage of first-place votes each method received across three evaluation dimensions. A total of 58 participants provided valid rankings. The best result for each dimension is highlighted in bold.
  }
  \label{tab_app:user}
  \vspace{5pt}
  \begin{tabular*}{.65\textwidth}{l@{\extracolsep{\fill}}ccc} 
    \toprule
    Evaluation Dimension & IN4D & CTRL-D & Ours \\
    \midrule
    Naturalness             & $31.72\%$ & $28.62\%$ & $\textbf{39.66}\%$  \\
    Prompt Fidelity         & $27.24\%$ & $21.03\%$ & $\textbf{51.72}\%$  \\
    Background Preservation & $26.90\%$ & $35.52\%$ & $\textbf{37.59}\%$  \\
    \bottomrule
  \end{tabular*}
\end{table}

\subsection{More Editing Results}
\label{sec_app:more_editing_results}
\begin{figure*}[t]
  \centering
  \includegraphics[width=\textwidth]{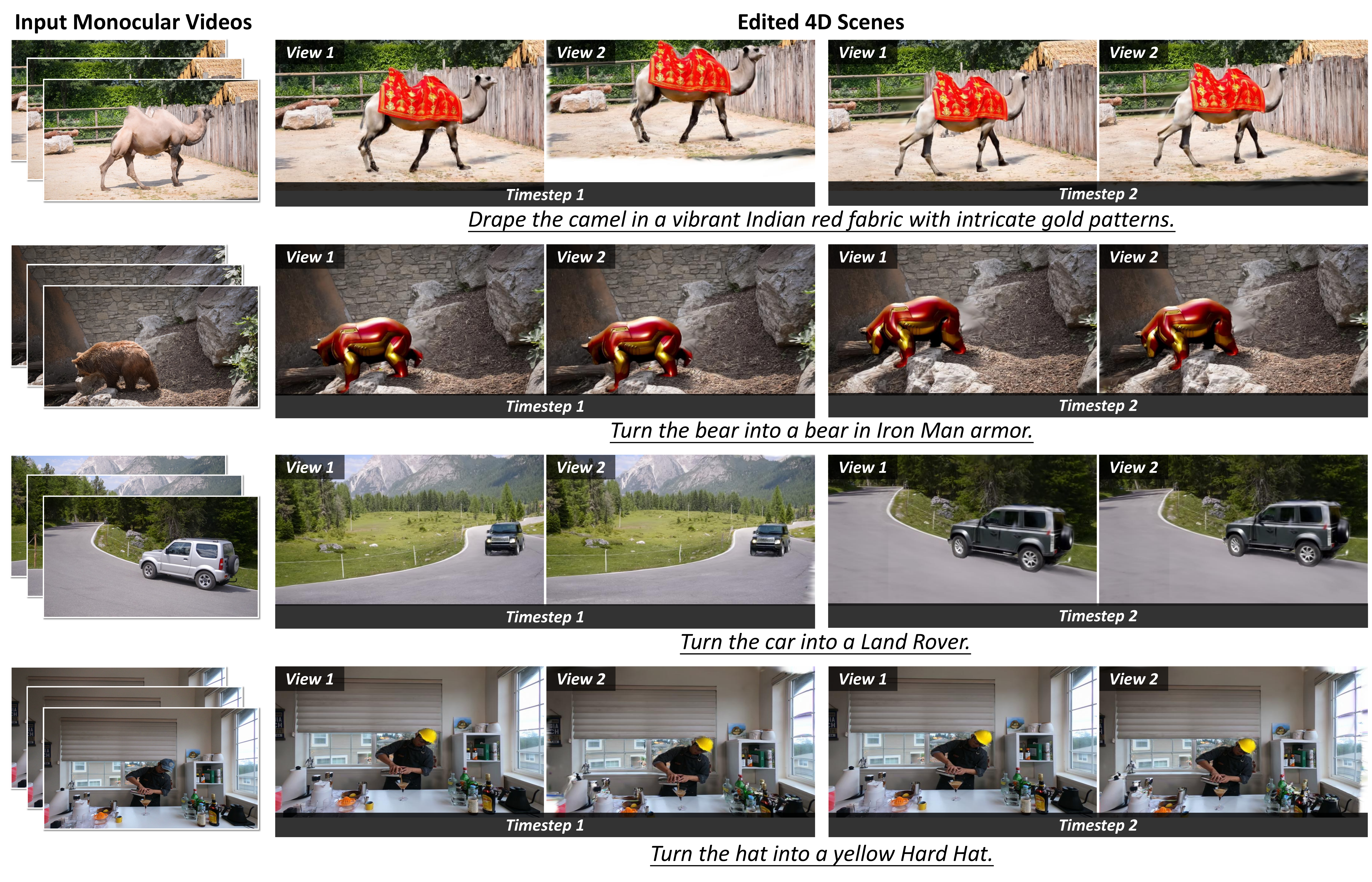}
  \caption{
    Editing results of \sysname{} on DAVIS and DyNerf datasets. 
    Each example shows monocular video input and the text-driven edited output at two different time steps, rendered from two novel views.
    Our method achieves accurate localization, realistic appearance changes, and preserves spatial and temporal consistency across views.
  }
  \label{fig:editing_results}
\end{figure*}

\begin{figure*}[t]
  \centering
  \includegraphics[width=\textwidth]{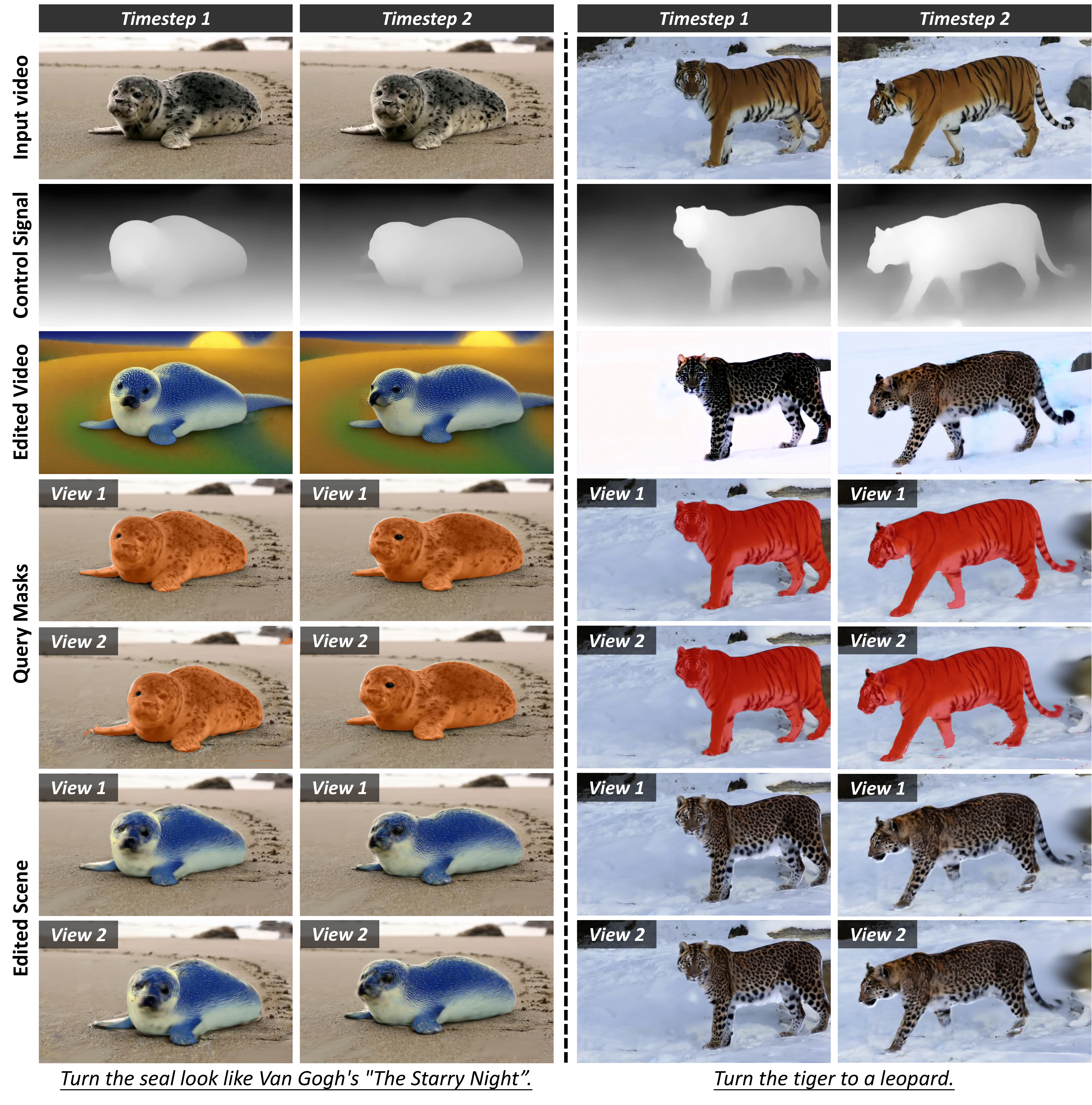}
  \caption{
    Editing results of \sysname{} on in-the-wild videos. Each example shows monocular video input, control signal, edited video and query masks and the text-driven edited output at two different time steps, rendered from two novel views. Our method achieves accurate localization, realistic appearance changes and preserves the background in its original state.
  }
  \label{fig:editing_results_w_video}
\end{figure*}

\begin{figure*}[t]
  \centering
  \includegraphics[width=\textwidth]{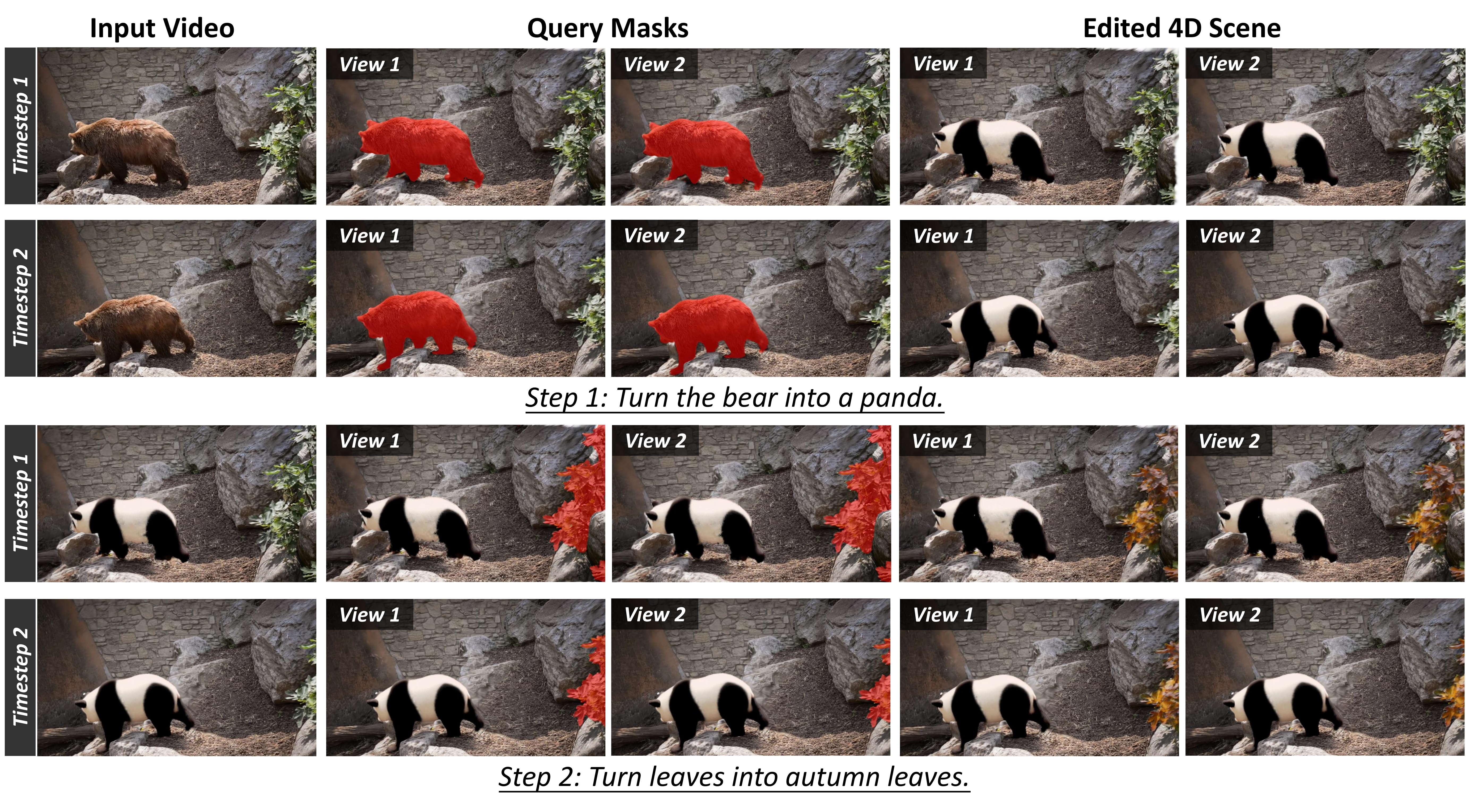}
  \caption{
    Multi-stage editing results of \sysname{} on DAVIS datasets. \sysname{} edits the scene in two stage: turn the bear into a panda in Step 1 and turn leaves into autumn leaves in Step 2. Each step shows query masks and the text-driven edited output at two different time steps, rendered from two novel views. Our method enables stage-by-stage editing of scenes while ensuring high-quality editing at each stage.
  }
  \label{fig:editing_separate}
\end{figure*}

To better demonstrate the effectiveness of our editing capabilities, we have conducted additional editing experiments on the Davis, DyNeRF datasets, and some videos collected from wild (Figure~\ref{fig:editing_results}, Figure~\ref{fig:editing_results_w_video}, Figure~\ref{fig:editing_separate}). In Figure~\ref{fig:editing_results}, we present the input videos from the DAVIS and DyNeRF datasets, along with the rendered images generated from the text-driven editing results in different timestamps and viewpoints. In Figure ~\ref{fig:editing_results_w_video}, we show the input image, the control signals of the video editing model, and the output results of the video editing model. Additionally, we render the relevance maps derived from text queries and the edited images from edited 4D scenes at different viewing angles and time points. In Figure~\ref{fig:editing_separate}, to demonstrate the multistage editing capability, we perform text-driven editing on different parts of the scene at each stage. We present the input video, along with rendered relevance maps corresponding to text queries and edited results from different viewpoints at various timestamps for each stage. 

These results indicate that our method effectively localizes the regions to be edited according to text instructions and produces high-quality edits with strong spatiotemporal consistency. Moreover, our approach also prevents the video model from inadvertently modifying the background, ensuring that modifications are restricted to the intended areas.

\begin{figure*}[t]
  \centering
  \includegraphics[width=\textwidth]{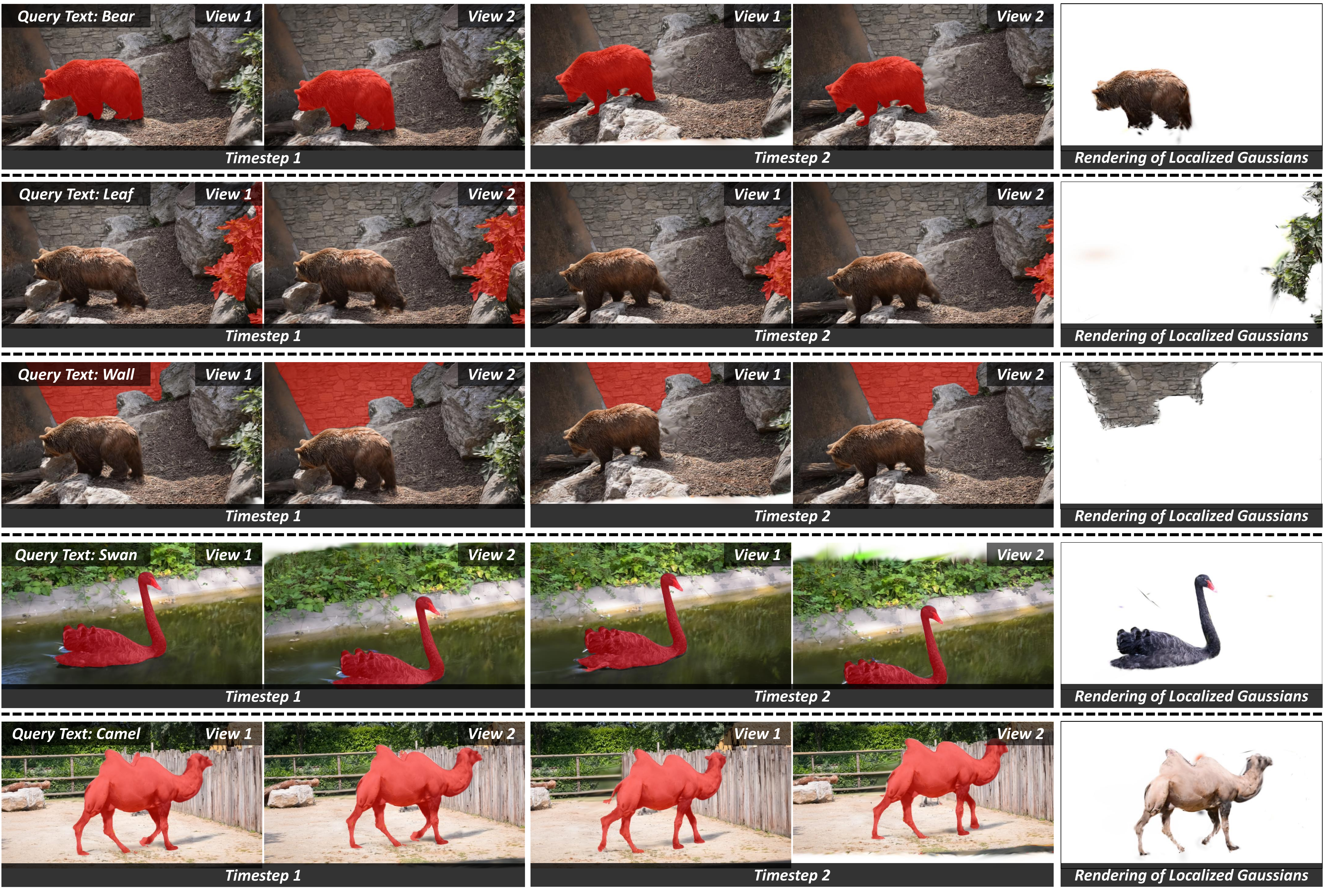}
  \caption{
    Visualization of 2D semantic queries and point-level localization.
    For each scene, we show the 2D relevance map generated from the language-embedded Gaussian features from the original and novel views at multiple time steps, as well as the rendering results of localized Gaussians.
    Our method accurately isolates both static and dynamic regions relevant to the input query.
  }
  \label{fig:localization}
\end{figure*}

\subsection{Semantic Querying and Localization Visualization}
\label{sec_app:semantic_query_localization}
To further analyze the semantic precision of our approach, we visualize the 2D query maps and point-level Gaussian localizations in Figure~\ref{fig:localization}. We render 2D relevance maps by querying the language-embedded Gaussian features with text prompts in both the original and novel views at multiple time steps, and visualize the localized Gaussians.

These results demonstrate that our Language-Embedded Dynamic Gaussians capture rich semantic information and enable fine-grained localization of both static and dynamic scene elements. For instance, the method successfully isolates moving subjects such as an animal while excluding background content, ensuring that only intended regions are subject to editing.

\subsection{Ablation Study of Video Editing Model}
\label{sec_app:ablation_of_video}
\begin{figure*}[t]
  \centering
  \includegraphics[width=\textwidth]{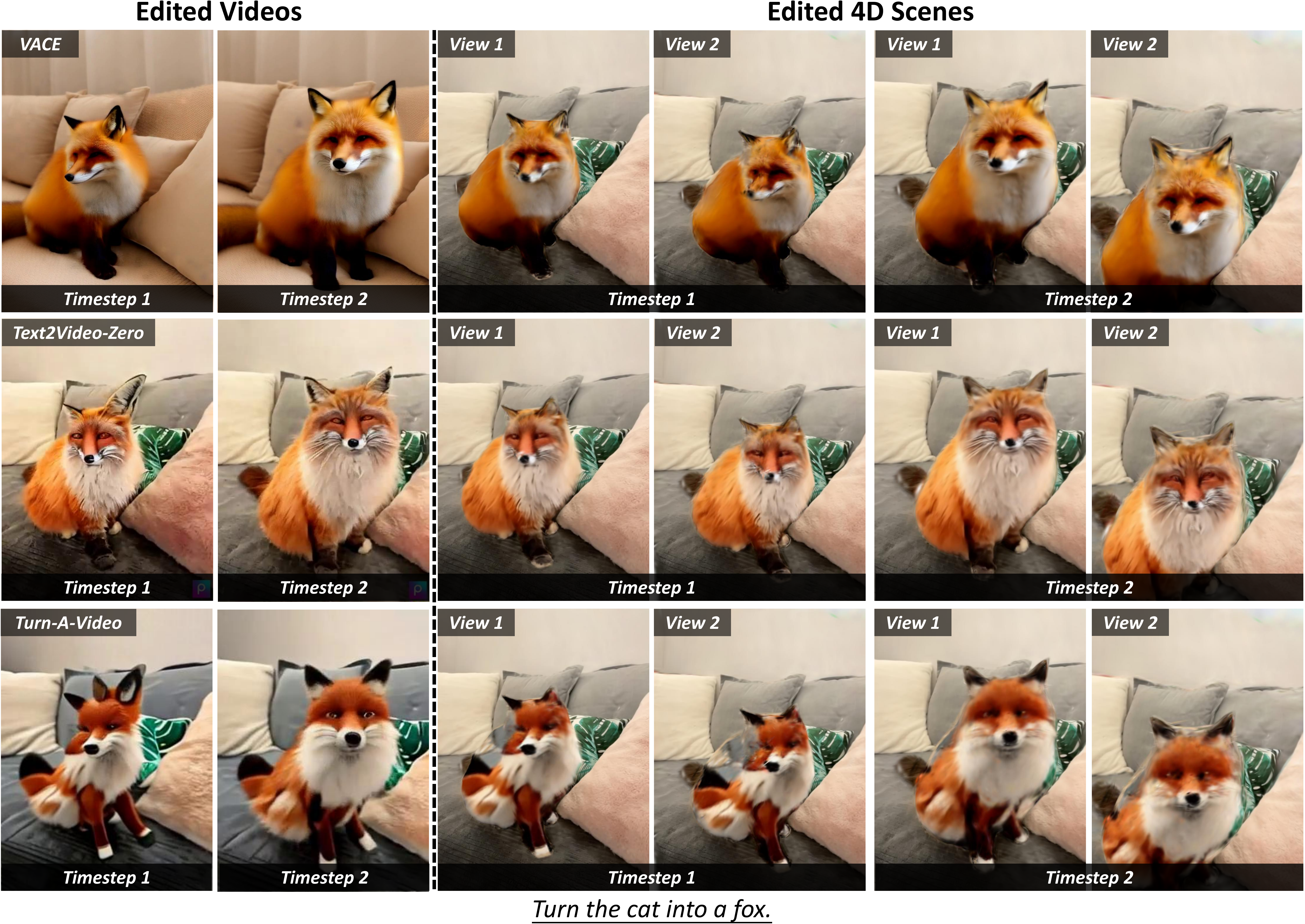}
  \caption{
    Qualitative ablation study on the effect of different video editing methods, including Edited Videos results and Edited 4D Scenes results using (1) VACE~\citep{jiang2025vace}, (2) Text2Video-Zero~\citep{khachatryan2023text2video} and (3) Tune-A-Video~\citep{wu2023tune}. 
    VACE shows natural and consistent effects across different timesteps and views, maintaining temporal coherence in the edited region. Text2Video-Zero causes unnatural facial features and fur textures, while Tune-A-Video fails to preserve the original shape and motion features. Importantly, our method ensures that irrelevant areas of the 4D scenes remain unaffected, while VACE further excels by providing superior temporal and spatial consistency in the edited areas.}
  \label{fig:video_ablation}
\end{figure*}

Figure~\ref{fig:video_ablation} presents a comparison of editing results from different video editing models on the DyCheck iPhone dataset. Specifically, the result edited by Tune-A-Video~\citep{wu2023tune} fails to preserve the original motion features; the result edited by Text2Video-Zero~\citep{khachatryan2023text2video} shows unnatural visual effects; in contrast, the result edited by VACE not only well retains the original features but also achieves high-quality editing effects. 

Across different video editing models, our method consistently ensures that the regions unrelated to the edit remain unaffected, thereby preserving the original scene context. Among these models, VACE achieves the best overall performance: it not only provides high-quality editing results but also maintains strong temporal consistency in the edited regions, ensuring smooth and coherent transitions across frames. 

Notably, VACE alone fails to keep semantically irrelevant regions unchanged—only when integrated with our method can this limitation be resolved. For other less-performing models that may introduce flaws (e.g., temporal inconsistency, shape distortion) in edited regions, our method still effectively prevents semantically irrelevant areas from unintended alterations. 
This highlights both VACE’s strengths in high-fidelity, temporally coherent edits and the indispensability of our method. Our method not only complements VACE to protect non-edited regions but also generalizes across video editing models. This further validates the advantage of our framework for stable, natural 4D scene edits.

\end{document}